\documentclass[runningheads]{llncs}
\usepackage{graphicx}
\usepackage{multirow}
\usepackage{booktabs}
\usepackage{cite}
\usepackage{enumitem}
\usepackage{amsmath,graphicx}
\usepackage{multicol}
\usepackage{rotating} 
\begin{document}
\title{Improve Convolutional Neural Network Pruning by Maximizing Filter Variety}
\titlerunning{Improve CNN Pruning by Maximizing Filter Variety}

\author{Nathan Hubens\inst{1,2,}\thanks{This research has been conducted in the context of a joint-PhD between the two institutions.} \and
Matei Mancas\inst{1} \and
Bernard Gosselin \inst{1} \and
Marius Preda \inst{2} \and
Titus Zaharia \inst{2}
}
\authorrunning{N. Hubens et al.}
%
\institute{ISIA, Faculty of Engineering of Mons, UMONS, Belgium
\email{\{firstname.lastname\}@umons.ac.be}
\and
ARTEMIS, Telecom SudParis, IP Paris, France \\
\email{\{firstname.lastname\}@telecom-sudparis.eu}\\
}

%

%
\maketitle 
\begin{abstract}



Neural network pruning is a widely used strategy for reducing model storage and computing requirements. It allows to lower the complexity of the network by introducing sparsity in the weights. Because taking advantage of sparse matrices is still challenging, pruning is often performed in a structured way, \textit{i.e.} removing entire convolution filters in the case of ConvNets, according to a chosen pruning criteria. Common pruning criteria, such as $l_1$-norm or movement, usually do not consider the individual utility of filters, which may lead to: (1) the removal of filters exhibiting rare, thus important and discriminative behaviour, and (2) the retaining of filters with redundant information. In this paper, we present a technique solving those two issues, and which can be appended to any pruning criteria. This technique ensures that the criteria of selection focuses on redundant filters, while retaining the rare ones, thus maximizing the variety of remaining filters. The experimental results, carried out on different datasets (CIFAR-10, CIFAR-100 and CALTECH-101) and using different architectures (VGG-16 and ResNet-18) demonstrate that it is possible to achieve similar sparsity levels while maintaining a higher performance when appending our filter selection technique to pruning criteria. Moreover, we assess the quality of the found sparse subnetworks by applying the Lottery Ticket Hypothesis and find that the addition of our method allows to discover better performing tickets in most cases.

\end{abstract}

\section{Introduction}

Convolutional Neural Networks have been applied to a wide variety of computer vision tasks and have exhibited state-of-the-art results in most of them. Their recent success was partially due to an increase in their complexity and depth at the expense of increasing the needs of parameter storage and computation. This drawback makes it challenging for deep neural networks to be used for applications with limitations in terms of memory and/or processing time, such as embedded systems or real-time applications. \\

Recent studies have exhibited a particular characteristic of neural networks, called the Lottery Ticket Hypothesis \cite{lth}. This hypothesis suggests that, in regular neural network architectures, there exists a subnetwork that can be trained to the same level of performance as the original one, as long as it starts from the same original conditions. This implies that an important reason why complex architectures are successful nowadays is because, by possessing many parameters, they have more chance to contain such a ``winning ticket". At the same time, once a winning ticket has been discovered, all the other parameters can be removed without affecting the model's performance. The technique consisting of removing unnecessary parameters and inducing sparsity in a neural network is called \textit{neural network pruning}. \\

To prune a neural network, the most commonly used criteria of selection is the magnitude pruning, also called $l_1$ pruning, \textit{i.e.} removing parameters having the lowest absolute value. Even though it allows to reach non-trivial sparsity levels, magnitude pruning still presents two main shortcomings. First, while being very efficient when the network’s weights have been randomly initialized, magnitude pruning has shown limitations in the transfer-learning regime, \textit{i.e.} when the network’s weights come from pre-training on a larger dataset. Indeed, in the transfer learning regime, final weight values are mostly predetermined by the original model \cite{mvmt}. Thus, high magnitude weights that were useful for the pre-training task are not necessarily useful for the new end task. This was recently solved by the movement pruning criteria \cite{mvmt}, which removes weights whose value moves towards zero. Second, magnitude pruning does not explicitly seek to remove redundant parameters and to maximize the variety of filters that the network contains after pruning. Methods attempting to increase the variety of filters by grouping them by similar functionality have shown promising results \cite{fop}. \\

In this work we propose to slightly alter the usual pruning process and more particularly the filter selection mechanism by introducing a clustering method, ensuring that parameters exhibiting similar/redundant behaviours can be pruned while those with unique behaviour are being kept. The contributions of this paper are the following:
\begin{itemize}
    \item Propose a simple and intuitive algorithm grouping redundant features and ensuring that filters extracting unique features are not removed in the pruning process. This algorithm may be appended to any filter selection criteria.
    \item Empirically show that the proposed method improves storage and computation costs over recently proposed techniques across several benchmark models and datasets without affecting the network performance.
    \item Show that using the proposed method, we are able to discover better subnetworks, following the Lottery Ticket Hypothesis.
\end{itemize}

\section{Related Work}
\label{rw}

Pruning techniques can differ in many aspects. The three main axis of differentiation are detailed in this section. \\

\textbf{Granularity. } When pruning a neural network, it is first required to define the granularity, \textit{i.e.} the structure according to which parameters will be removed. Granularities of pruning are usually categorized into two groups: \textit{unstructured} pruning, \textit{i.e.} when individual weights are evaluated and removed \cite{obd, obs}. This leads to sparse weight matrices, often difficult to optimize in terms of computation and speed efficiency. For those reasons, \textit{structured} pruning was introduced. This kind of pruning takes care of removing blocks of weights, which can take the form of vectors, kernels or even filters \cite{pfec, cpa, hub}. While most of structured pruning granularities still leaves sparse weight matrices, as the weights are removed in blocks, it makes it easier for dedicated hardware or libraries to take advantage of the removal of those weights. Filter pruning is a particular case as sparse filters can just simply be removed from the network architecture, thus keeping the network dense, and not requiring any specialized sparse computation library to obtain computation speed-up and storage reduction. \\

\textbf{Criteria. } In order to know which weights, or group of weights, will be pruned, we need to evaluate their importance. For that purpose, we define a criteria, ranking each parameter, and remove those that have the lowest score. Early work made use of second-order approximation of the loss surface to select the parameters to remove \cite{obd, obs}. Due to the important computation overhead that such a technique introduces, other criteria have emerged, such as $l_1$-norm, $l_2$-norm or first-order Taylor expansion. Other training techniques enforcing sparse structures such as $l_0$ regularization \cite{l0} or variational dropout \cite{var_dropout} have also been introduced. To this day, the criteria of selection which is the most commonly used due to its simplicity yet providing good and generalizable results across many datasets and architectures is the magnitude pruning \cite{state}, based on $l_1$-norm, the weights with the lowest absolute value assumed to be the least important, as they will produce the weakest activations and thus, participate the least to the output of the neural network. Moreover, the criteria may be used to compare weights belonging to a common layer, \textit{i.e.} local pruning, thus leading to a structure with layers of equivalent sparsities. The criteria may also be applied to the whole network, comparing weights from all layers, \textit{i.e.} global pruning, and leading to a network with layers of different sparsity levels. While global pruning possesses a more computation overhead since it potentially compares millions of parameter at each pruning step, it usually provides better results as it allows for more freedom of parameter selection. \\

\textbf{Scheduling. } The pruning scheduling defines how pruning is integrated in the training process. Early pruning methods cared about removing redundant weights after the network has been trained to convergence, and performing the pruning in a single-step, which is nowadays called \textit{one-shot pruning} \cite{pfec}. It was soon discovered that \textit{iterative pruning}, performing the pruning in several steps, alternating pruning and fine-tuning of the network to allow it to recover from the lost performance, helped to reach more extreme sparsity levels and to let the network to more easily recover the pruning of many weights \cite{taylor, han}. Recently, research about the Lottery Ticket Hypothesis have shown that the optimal pruned network could be discovered from the very initial state of a neural network, \textit{i.e.} before any training has occurred \cite{lth}. While it is still very difficult to uncover such an optimal network right from initialization, it inspired many research to prune the network earlier in the training process, \textit{i.e.} not starting from a first pre-training phase, but also to propose more complex scheduling functions which integrates pruning early in the training process \cite{ocp, to_prune}. \\

In our work, we focus on a particular structured pruning approach, filter pruning. More particularly, our contribution concerns the criteria of selection of filters. We propose a slight alteration that can be inserted into the iterative pruning process, allowing to use any state-of-the-art criteria, but helping them to select redundant filters to remove, while preserving unusual, thus potentially discriminative filters.

\section{Proposed Methodology}
\label{Methodology}

\begin{figure}[!htbp]
\centering
\includegraphics[width=\textwidth]{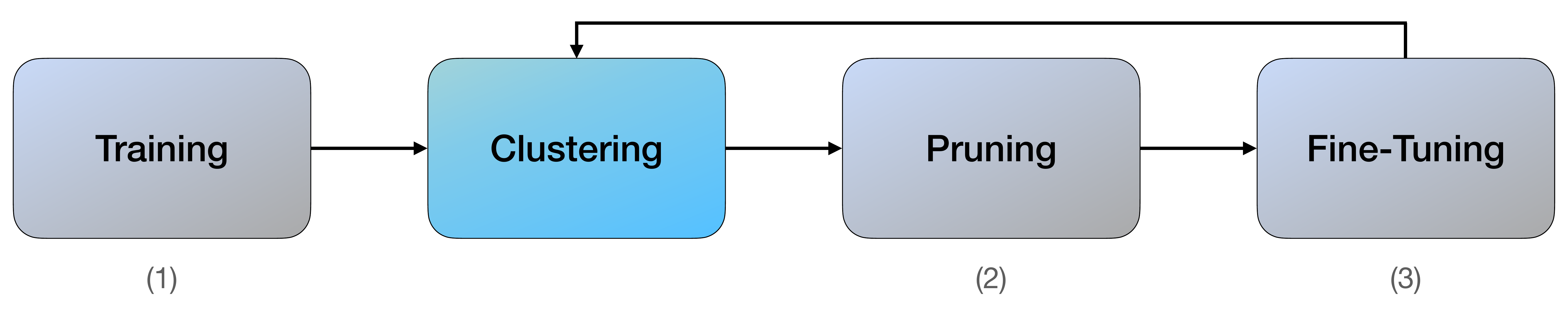}
\caption{The proposed pruning pipeline. We introduce a fourth step in the common iterative pruning process, aiming to cluster convolution filters by similar functionality. By then performing pruning in each cluster, we ensure that we remove redundant filters, while preserving the rare ones that could have been removed otherwise.}
\label{process}
\end{figure}

Performing filter pruning following an iterative schedule usually consists of a three-step method, represented in Figure \ref{process}: (1) train the network to convergence, (2) prune a portion of the filters, according to a chosen criterion and, (3) fine-tune so that the model can recover from the lost performance. Steps (2) and (3) are then repeated, alternating pruning and fine-tuning until the desired sparsity is reached. We propose to add an additional step between (1) and (2). Indeed, before selecting the weights to remove according to a chosen criteria, we first would like to cluster filters exhibiting similar behaviour and to perform pruning in each cluster separately, and consequently only on redundant filters. By doing so, pruning will only retain independent filters while also retaining filters which have uncommon behaviours, thus maximizing the variety of remaining filters.

To learn about the functionality of each filter, we make use of a neural network interpretation technique named \textit{Activation Maximization} \cite{feat_viz1}. This method uses the gradient ascent optimization technique, starting from a random noise image, to modify each pixel of the image in order to maximize the activation of a particular convolution filter. In other words, the synthesized image is the image of features that excites the most a selected filter and thus, the feature it is the most sensitive to when processing a natural image. \\

For each convolutional filter in a layer of the network, we can synthesize its corresponding ``signature" image based on Activation Maximization. The goal is then to perform K-Means clustering of those images, effectively grouping similar images together while also keeping unique ones into their dedicated group. To effectively reduce dimensionality and facilitate the task of K-Means clustering, we first feed our images to the convolutional part of an AlexNet model \cite{alexnet} pretrained on ImageNet \cite{imagenet}, encoding those into a feature vector, which will serve as input data to the clustering algorithm. \\

\begin{figure}[!htbp]
\centering
\includegraphics[width=\textwidth]{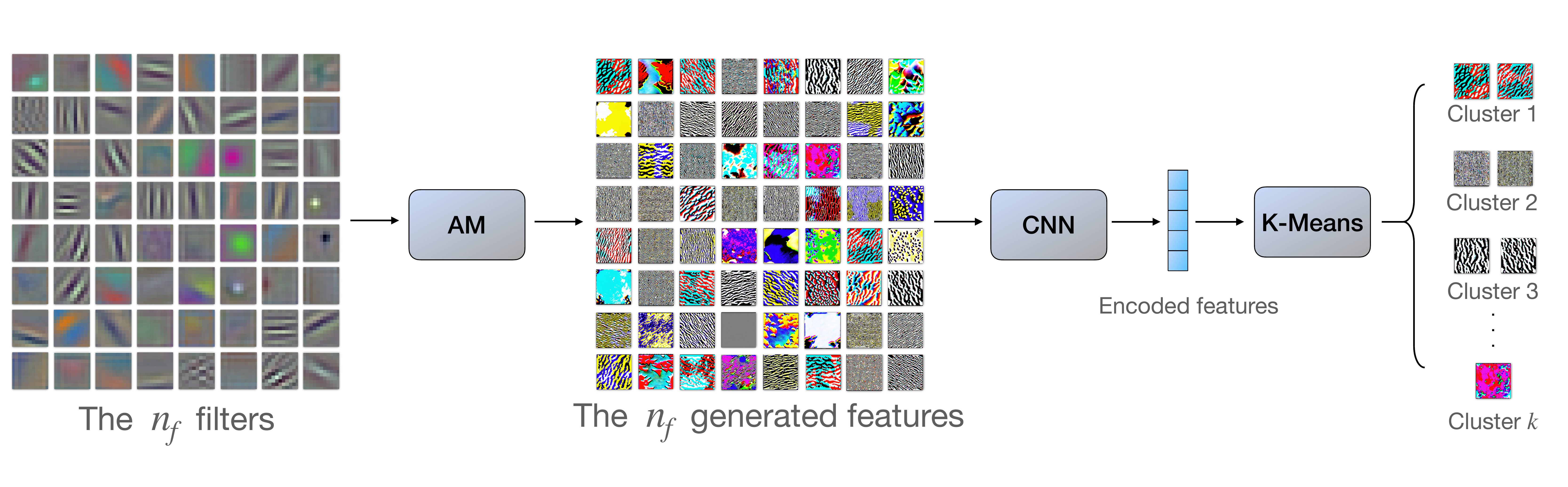}
\caption{Representation of the clustering process. We first extract each filter of a given layer, then generate the corresponding feature images with Activation Maximization technique. Those synthesized images are then encoded to a lower dimension by a pre-trained ConvNet, and clustered with K-Means. This technique allows to group filters sensitive to similar features together.}
\label{pipeline}
\end{figure}

Once each feature image has been clustered, we can then apply the pruning process, selecting remaining filters according to a chosen criteria but, this time not by comparing all the filters in the layer, but by comparing filters whose feature images are located in the same group. By doing so in each group, we will then only preserve the single best representative filter of each feature. The number of clusters $k$ is thus chosen as a compression parameter, depending on the desired sparsity. By setting a high number of clusters $k$, this creates more groups and thus removes less filters. \\

\begin{figure}[!h]
\centering
\includegraphics[width=0.8\textwidth]{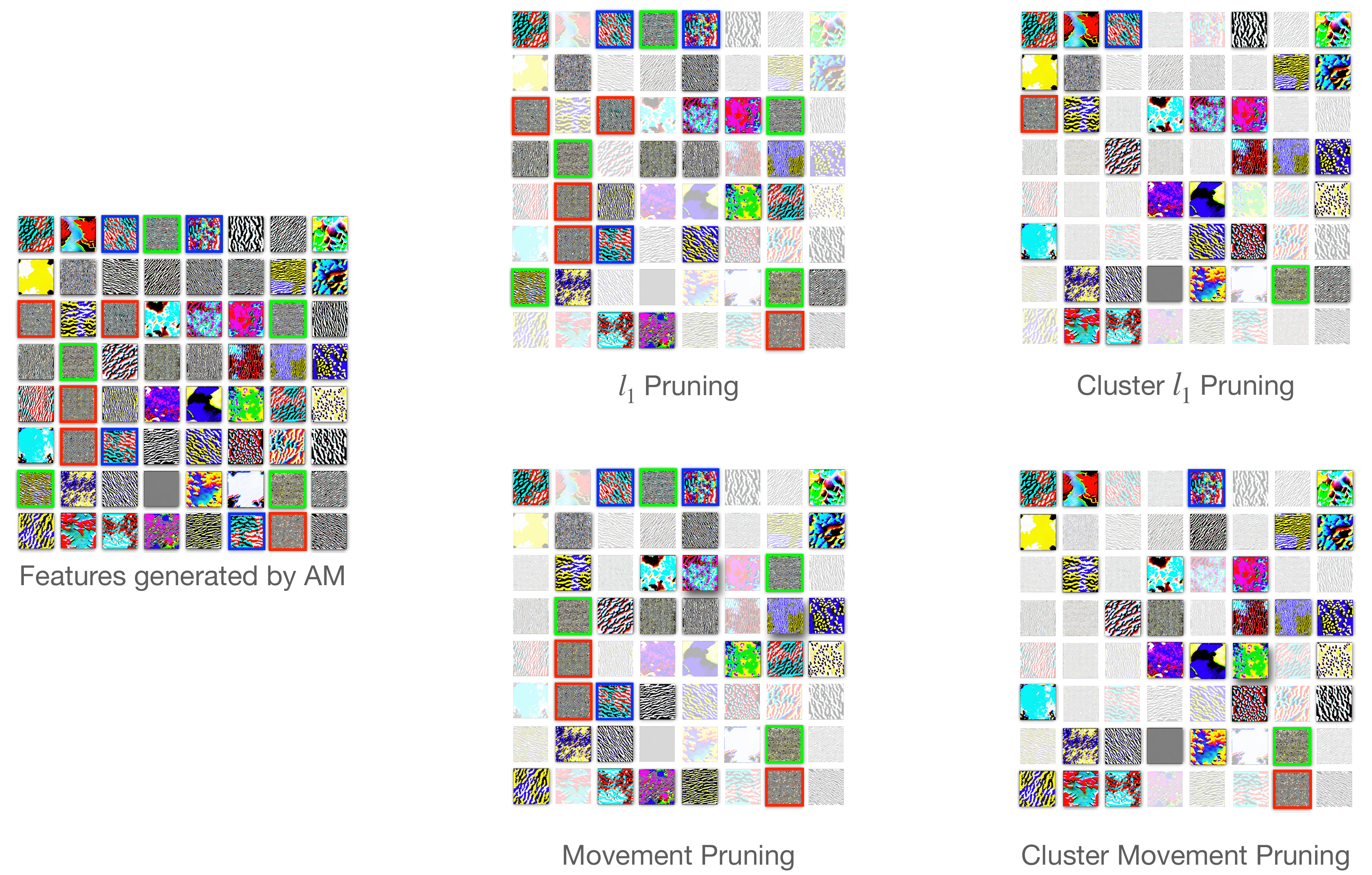}
\caption{Comparison of the remaining features after applying different pruning techniques until a sparsity of $50\%$ in the first layer of AlexNet. Three dominant clusters are highlighted in color. Features removed by pruning are greyed out.}
\label{clusters}
\end{figure}

When comparing common criteria before and after the addition of our clustering method, we observe that a greater variety of filters are retained. As an example, Figure \ref{clusters} represents features extracted from the filters of the first layer of a simple ConvNet, AlexNet \cite{alexnet}, by using the Activation Maximization technique. Three clusters of similar features have been highlighted in color, and the corresponding remaining features are shown for each pruning technique, with removed one being greyed out. We can observe that, by clustering similar features, we ensure that: (1) redundant features are removed and (2) rare features are being kept, which is not the case when using magnitude and movement pruning alone, where some features belonging to a same cluster are still present.

\section{Comparison to Common Criteria}
\label{Experiments}

In this section, we evaluate the effects of adding the extra clustering step presented in section \ref{Methodology} in the iterative pruning pipeline and apply the pruning criteria in each group separately. \\

\textbf{Datasets and Architectures.} For our experiments, the datasets have been chosen to be various in terms of image resolution and number of classes. In particular, we evaluate our methods on the three following datasets: (1) CIFAR-10 \cite{cifar}, composed of RGB images distributed in 10 classes, and of resolution $32\times 32$. (2) CIFAR-100 \cite{cifar}, composed of RGB training images distributed in 100 classes and of resolution $32 \times 32$ and (3) Caltech-101 \cite{caltech}, composed of pictures of objects, distributed in 101 classes and of resolution $300 \times 200$. Those datasets are then tested on two types of popular convolutional network architectures: VGG-16 \cite{vgg} and ResNet-18 \cite{resnet}. In particular, we use a modified version of VGG-16 which consists of 13 convolutional layers and 2 fully-connected layers, with each convolutional layer being followed by a batch normalization layer \cite{vgg_bis}. ResNet-18 belongs to a family of ConvNets using residual connections \cite{resnet}, and contains 17 convolutional layers and a single fully-connected layer. \\

\textbf{Training Procedure.} The networks we use for our experiments are initialized from pre-trained weights, \textit{i.e.} networks were previously trained on ImageNet and we reuse their weights. Images of our dataset are first resized to $224\times 224$ and are augmented by using horizontal flips, rotations, image warping and random cropping, then aggregated in batches of size $64$. We train each model using the 1cycle learning rate method \cite{1cycle}, where the training starts with a learning rate warmup until a nominal value, then gradually decay until the end of the training. \\

\textbf{Pruning Method.} We use the 4-step schedule presented in Figure \ref{process}. We propose to iterate over each layer and remove a specific amount of filters in order to reach the desired sparsity. As each layer will eventually have the same sparsity level, it can be considered as a form of local pruning. The first step is performed for $15$ epochs, at a learning rate of $1e-3$. Our clustering method then takes care of grouping the filters of a target layer into $k$ groups, $k=s \times n_f$, $s$ being the desired sparsity in percent and $n_f$ the amount of filters in that layer. After performing the pruning of filters and only keeping a single filter from each cluster, we fine-tune our model for $3$ epochs, with a learning rate of $3e-4$, to allow the network to recover from the loss of its parameters. This process is performed iteratively for each layer in the network. We evaluate the benefits of our extra step according to two pruning criteria: (1) $l_1$-norm of the filters, \textit{i.e.} remove the filters that possess the lowest norm, computed for each filter by $\sum_{i}|w_i|_t$ and (2) movement pruning, \textit{i.e.} only keep the filters whose magnitude has increased the most during training, computed by $\sum_{i}|w_i|_t - |w_i|_0$, with $|w_i|_t$, the weights values at training step $t$ and $|w_i|_0$, the weights values at initialization.\\

\textbf{Frameworks and Hardware. } These experiments are conducted using the PyTorch \cite{pytorch} and fastai \cite{fastai} libraries for the implementation of the training loop, fasterai \cite{fasterai} and Lucent \cite{lucent} for the implementation of the clustering and pruning methods and using a 12GB Nvidia GeForce GTX 1080 Ti GPU for computation.  \\

\textbf{Results. } The experiments, conducted on VGG-16 (Table \ref{table:vgg16}) and ResNet-18 (Table \ref{table:resnet18}), show that, for almost all sparsity levels, dataset and criteria tested, it is beneficial to the pruning process to add the proposed clustering step. Indeed, for the same sparsity level, accuracy increases up to $5\%$ can be observed, which also means that the same networks could be pruned to a higher sparsity level without witnessing performance degradation.

{\setlength{\tabcolsep}{0.4em}
\begin{table}[!htbp]
\begin{center}
\begin{tabular}{l|l||c c | c c}
\cmidrule[1pt]{3-6}
\multicolumn{2}{c}{}&\multicolumn{1}{c}{$l_1$ }&\multicolumn{1}{c}{Cluster $l_1$}&\multicolumn{1}{c}{Movement}&\multicolumn{1}{c}{Cluster Movement}\\\midrule
\multicolumn{2}{l}{\textbf{CIFAR-10}} & \multicolumn{3}{c}{}             &  \\\midrule
\multirow{3}{4mm}{\begin{sideways}\parbox{11mm}{Sparsity}\end{sideways}}
& 60\% & 90.89 $\pm$ 0.17 & \textbf{92.39 $\pm$ 0.07} & 91.55 $\pm$ 0.18 & \textbf{92.45 $\pm$ 0.26} \\

& 70\% & 89.47 $\pm$ 0.09 & \textbf{90.91 $\pm$ 0.13} & 90.35 $\pm$ 0.12  &  \textbf{90.89 $\pm$ 0.10}   \\

& 80\% & 84.95 $\pm$ 0.08  & \textbf{87.04 $\pm$ 0.23} & 86.50 $\pm$ 0.19 &  \textbf{87.48 $\pm$ 0.10}  \\\midrule

\multicolumn{2}{l}{\textbf{CIFAR-100}} & \multicolumn{3}{c}{}             &  \\\midrule
\multirow{3}{4mm}{\begin{sideways}\parbox{11mm}{Sparsity}\end{sideways}}
& 60\% & 54.94 $\pm$ 0.21 & \textbf{58.72 $\pm$ 0.30} & 54.81 $\pm$ 0.34 &  \textbf{57.29 $\pm$ 0.71} \\

& 70\% & 47.56 $\pm$ 0.94 & \textbf{51.67 $\pm$ 0.61} & 47.59 $\pm$ 0.29 &  \textbf{51.25 $\pm$ 0.41} \\

& 80\% & 35.67 $\pm$ 0.90 & \textbf{39.30 $\pm$ 0.65} & 36.82 $\pm$ 1.42 &  \textbf{42.30 $\pm$ 0.94}  \\\midrule

\multicolumn{2}{l}{\textbf{Caltech-101}} & \multicolumn{3}{c}{}             &  \\\midrule
\multirow{3}{4mm}{\begin{sideways}\parbox{11mm}{Sparsity}\end{sideways}}
& 60\% & 86.63 $\pm$ 0.39 & \textbf{87.44 $\pm$ 0.28} & 86.94 $\pm$ 0.28 &  \textbf{87.40 $\pm$ 0.51}  \\

& 70\% & 84.87 $\pm$ 0.41 & \textbf{86.01 $\pm$ 0.79} & 82.89 $\pm$ 0.19 &   \textbf{84.43 $\pm$ 0.25} \\

& 80\% & \textbf{79.18 $\pm$ 0.63}& 79.03 $\pm$ 0.72  & 76.00 $\pm$ 0.59 &  \textbf{78.75 $\pm$ 0.63}  \\\midrule

\end{tabular}
\end{center}
\caption{Results of applying different pruning criteria on VGG-16. The benefit of applying our clustering method before selecting the filters to remove translates to a higher accuracy for most sparsity levels and datasets. Values in bold are the best when comparing a single criteria with and without the clustering process. Accuracies and standard deviation over 3 runs are reported.}
\label{table:vgg16}
\end{table}}

{\setlength{\tabcolsep}{0.4em}
\begin{table}[!htbp]
\begin{center}
\begin{tabular}{l|l||c c | c c}
\cmidrule[1pt]{3-6}
\multicolumn{2}{c}{}&\multicolumn{1}{c}{$l_1$ }&\multicolumn{1}{c|}{Cluster $l_1$ }&\multicolumn{1}{c}{Movement}&\multicolumn{1}{c}{Cluster Movement}\\\midrule
\multicolumn{2}{l}{\textbf{CIFAR-10}} & \multicolumn{3}{c}{}             &  \\\midrule
\multirow{3}{4mm}{\begin{sideways}\parbox{11mm}{Sparsity}\end{sideways}}
& 60\% & 93.32 $\pm$ 0.11 & \textbf{93.76 $\pm$ 0.18} & 92.73 $\pm$ 0.16 &  \textbf{93.57 $\pm$ 0.10 }  \\

& 70\% & 92.17 $\pm$ 0.11 & \textbf{92.20 $\pm$ 0.11} & 90.79 $\pm$ 0.14 &  \textbf{92.35 $\pm$ 0.03}   \\

& 80\% & 89.58 $\pm$ 0.17 & \textbf{90.13 $\pm$ 0.09} & 87.04 $\pm$ 0.24 &   \textbf{89.53 $\pm$ 0.23 } \\\midrule

\multicolumn{2}{l}{\textbf{CIFAR-100}} & \multicolumn{3}{c}{}             &  \\\midrule
\multirow{3}{4mm}{\begin{sideways}\parbox{11mm}{Sparsity}\end{sideways}}
& 60\% & 71.65 $\pm$ 0.22 & \textbf{72.61 $\pm$ 0.41} & 70.95 $\pm$ 0.13 &  \textbf{72.27 $\pm$ 0.37 } \\

& 70\% & 67.18 $\pm$ 0.17 & \textbf{68.46 $\pm$ 0.25} & 66.19 $\pm$ 0.15 &  \textbf{68.44 $\pm$ 0.21}  \\

& 80\% & 59.14 $\pm$ 0.13 & \textbf{60.32 $\pm$ 0.30} & 58.50 $\pm$ 0.41 &  \textbf{59.86 $\pm$ 0.23 } \\\midrule

\multicolumn{2}{l}{\textbf{Caltech-101}} & \multicolumn{3}{c}{}             &  \\\midrule
\multirow{3}{4mm}{\begin{sideways}\parbox{11mm}{Sparsity}\end{sideways}}
& 60\% & 92.63 $\pm$ 0.05  & \textbf{93.00 $\pm$ 0.16}  & 90.77 $\pm$ 0.18  &  \textbf{91.81 $\pm$ 0.26}  \\

& 70\% & 88.75 $\pm$ 0.42 & \textbf{89.48 $\pm$ 0.30} & 85.32 $\pm$ 0.27 &  \textbf{87.75 $\pm$ 0.24}  \\

& 80\% & 79.89 $\pm$ 0.10 & \textbf{80.31 $\pm$ 0.35} & 75.05 $\pm$ 0.56 &  \textbf{76.29 $\pm$ 0.39}  \\\midrule

\end{tabular}
\end{center}
\caption{Results of applying different pruning criteria on ResNet-18. The benefit of applying our clustering method before selecting the filters to remove translates to a higher accuracy for most sparsity levels and datasets. Values in bold are the best when comparing a single criteria with and without the clustering process. Accuracies and standard deviation over 3 runs are reported.}
\label{table:resnet18}
\end{table}}

\section{Application to Lottery Ticket Hypothesis}

In order to not only compare the relative performance of the addition of our proposed clustering technique, we also would like to compare the quality of the remaining subnetworks, obtained after pruning. Such an analysis may be performed using the Lottery Ticket Hypothesis. \\

\textbf{Finding Lottery Tickets.} The Lottery Ticket Hypothesis states that: in each network, there exists a subnetwork that, trained in isolation, and from the same initial weight values, is able to achieve comparable performance in a comparable training time as the whole network. The authors proposed to find the said winning-tickets through an iterative process, repeatedly pruning a portion of remaining weights according to their $l_1$-norm, then resetting them to their initial value, \textit{i.e.} the value before any training has occured. Despite having shown promising results on small datasets, and for simple architectures, this method has shown difficulties to generalize to higher complexity use-cases \cite{lth}. To overcome this limitation, the same authors propose a slight weakening of the hypothesis. Instead of reinitializing the weights to their original value, they should be reinitialized to a value of an earlier step in the training process. This new hypothesis is called Lottery Ticket Hypothesis with Rewinding, and the sparse subnetworks are then called matching tickets instead of winning tickets \cite{lmc}.\\

\begin{figure}[h]
\begin{minipage}[b]{.49\linewidth}
  \centering
  \centerline{\includegraphics[width=5.8cm]{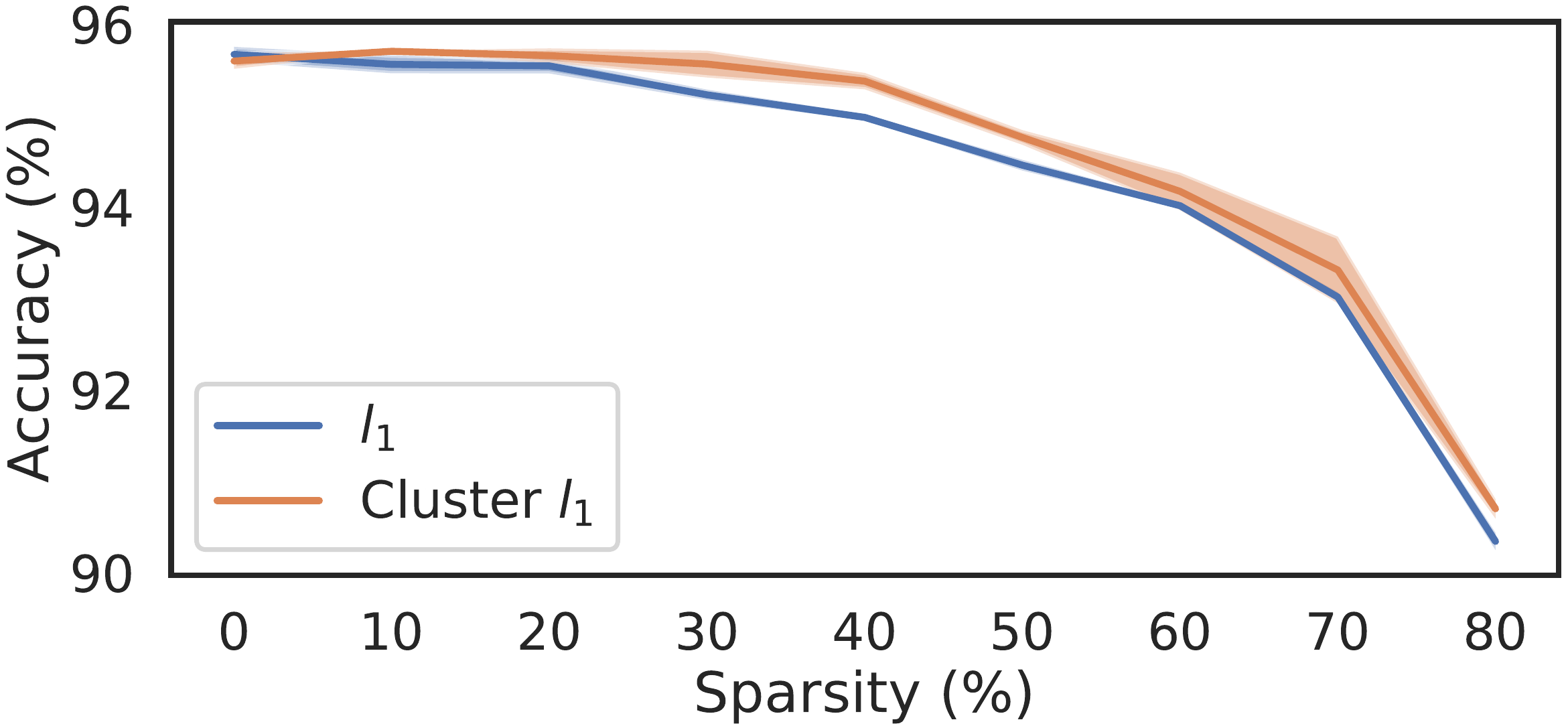}}
\end{minipage}
\hfill
\begin{minipage}[b]{0.49\linewidth}
  \centering
  \centerline{\includegraphics[width=5.8cm]{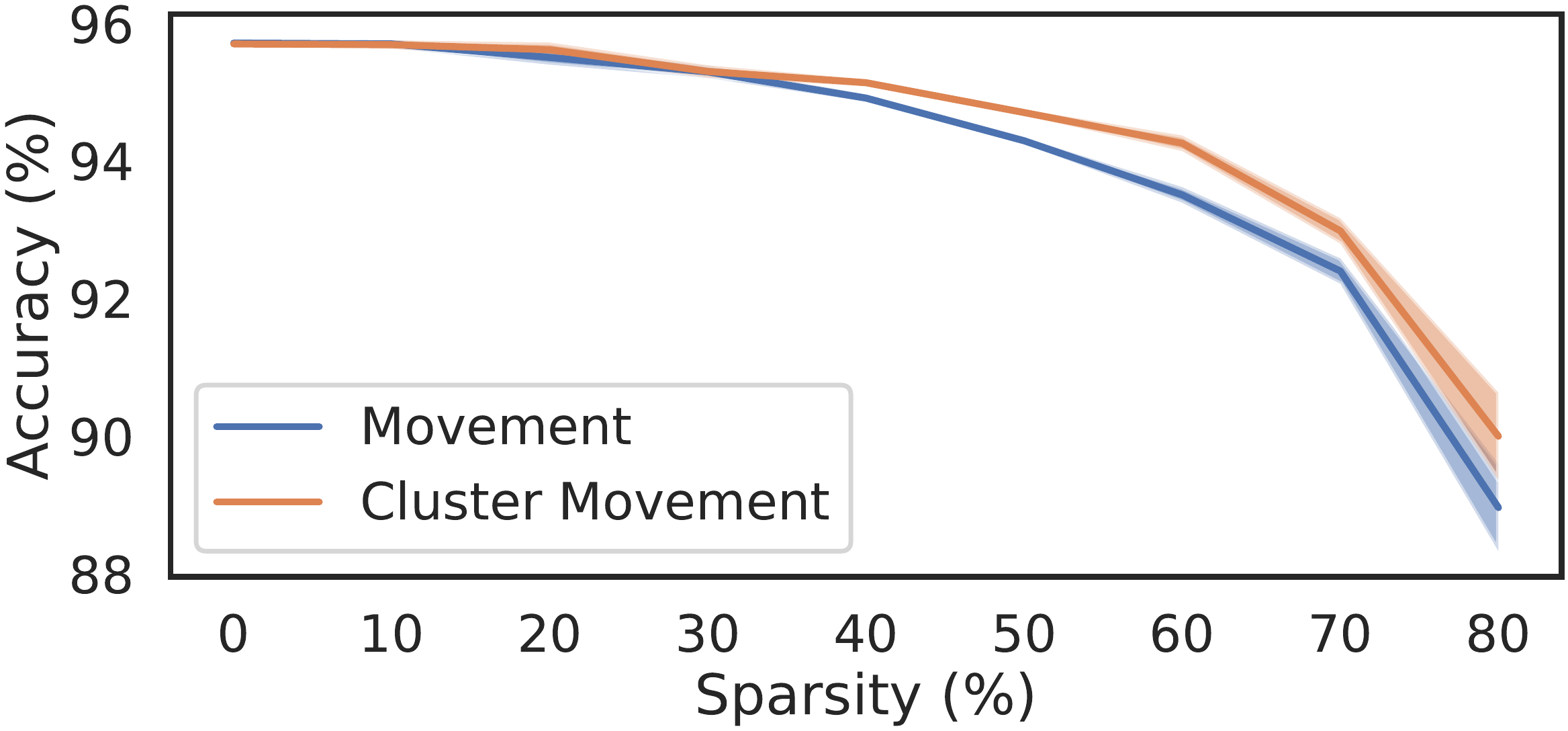}}
   \end{minipage}
      \caption{Results of the Lottery Ticket Hypothesis with Rewind test for different sparsities, performed with ResNet-18 on CIFAR-10.}
   
\begin{minipage}[b]{.49\linewidth}
  \centering
  \centerline{\includegraphics[width=5.8cm]{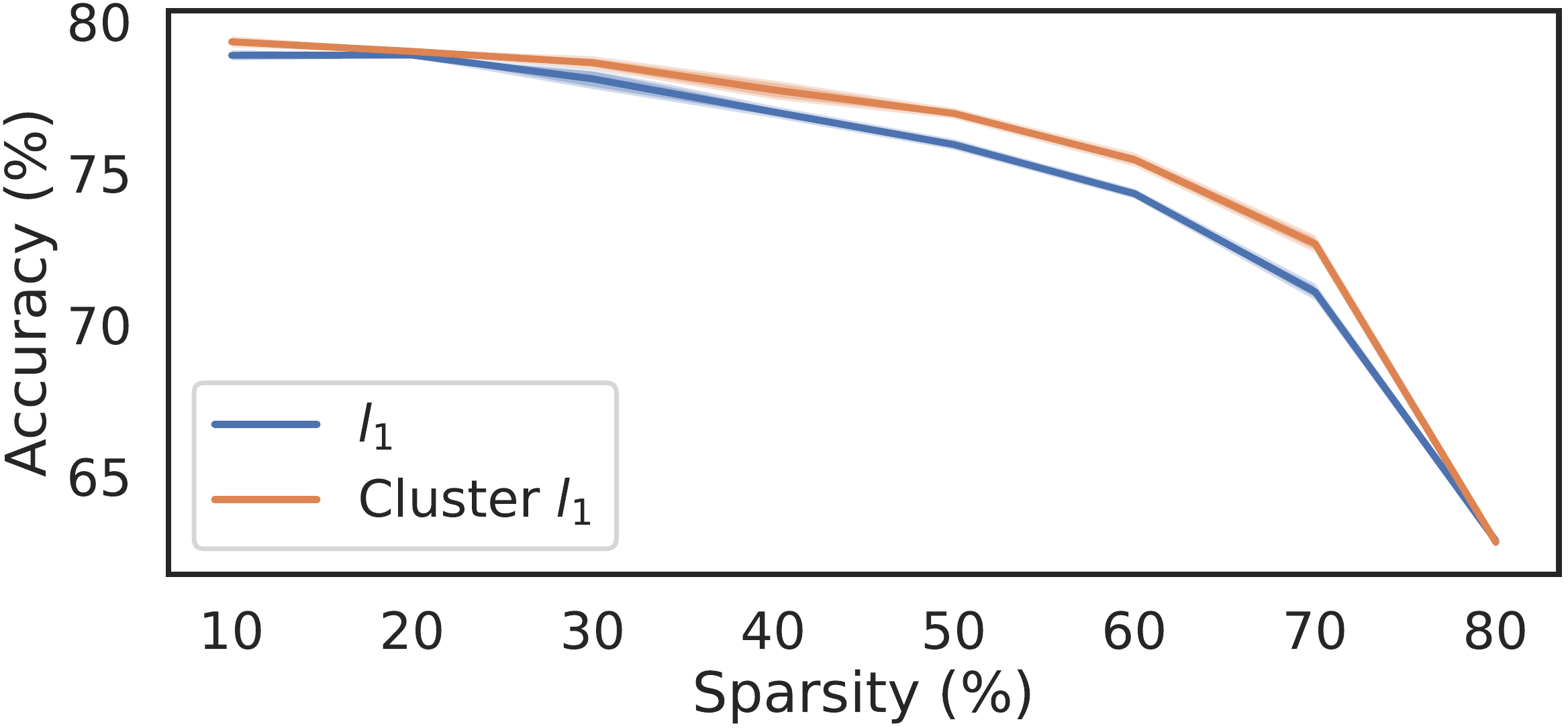}}
\end{minipage}
\hfill
\begin{minipage}[b]{0.49\linewidth}
  \centering
  \centerline{\includegraphics[width=5.8cm]{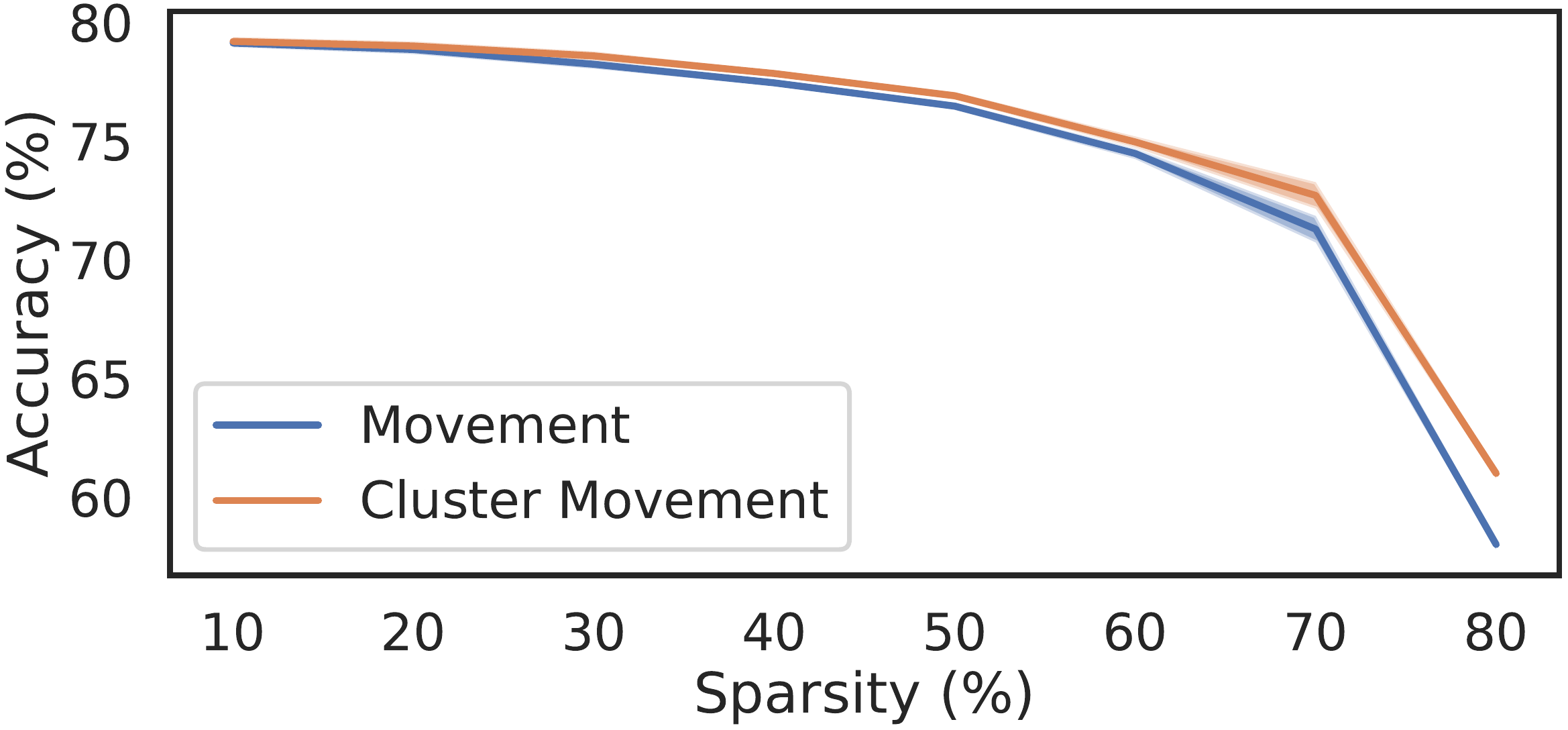}}
   \end{minipage}
   \caption{Results of the Lottery Ticket Hypothesis with Rewind test for different sparsities, performed with ResNet-18 on CIFAR-100.}
   
\begin{minipage}[b]{.49\linewidth}
  \centering
  \centerline{\includegraphics[width=5.8cm]{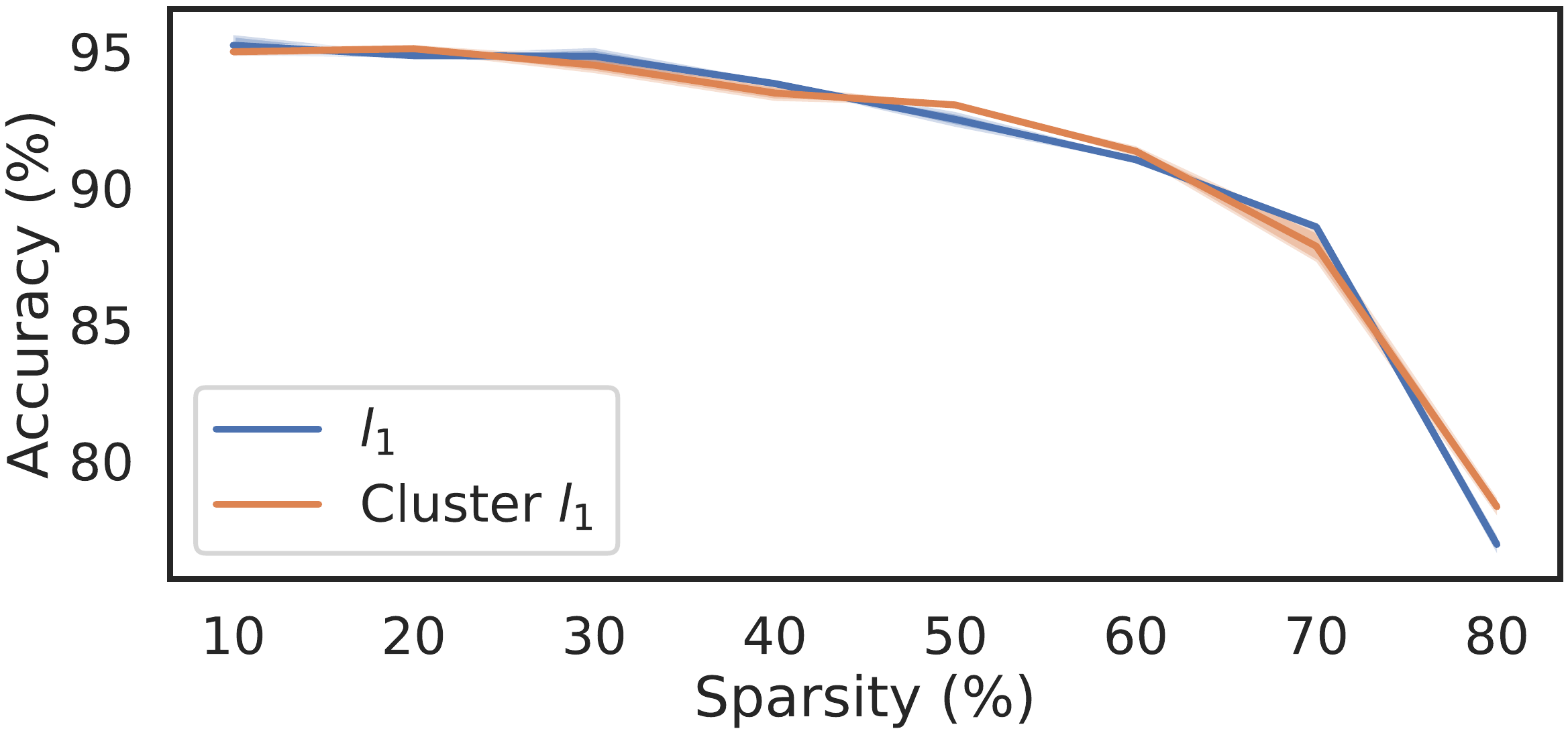}}
\end{minipage}
\hfill
\begin{minipage}[b]{0.49\linewidth}
  \centering
  \centerline{\includegraphics[width=6cm]{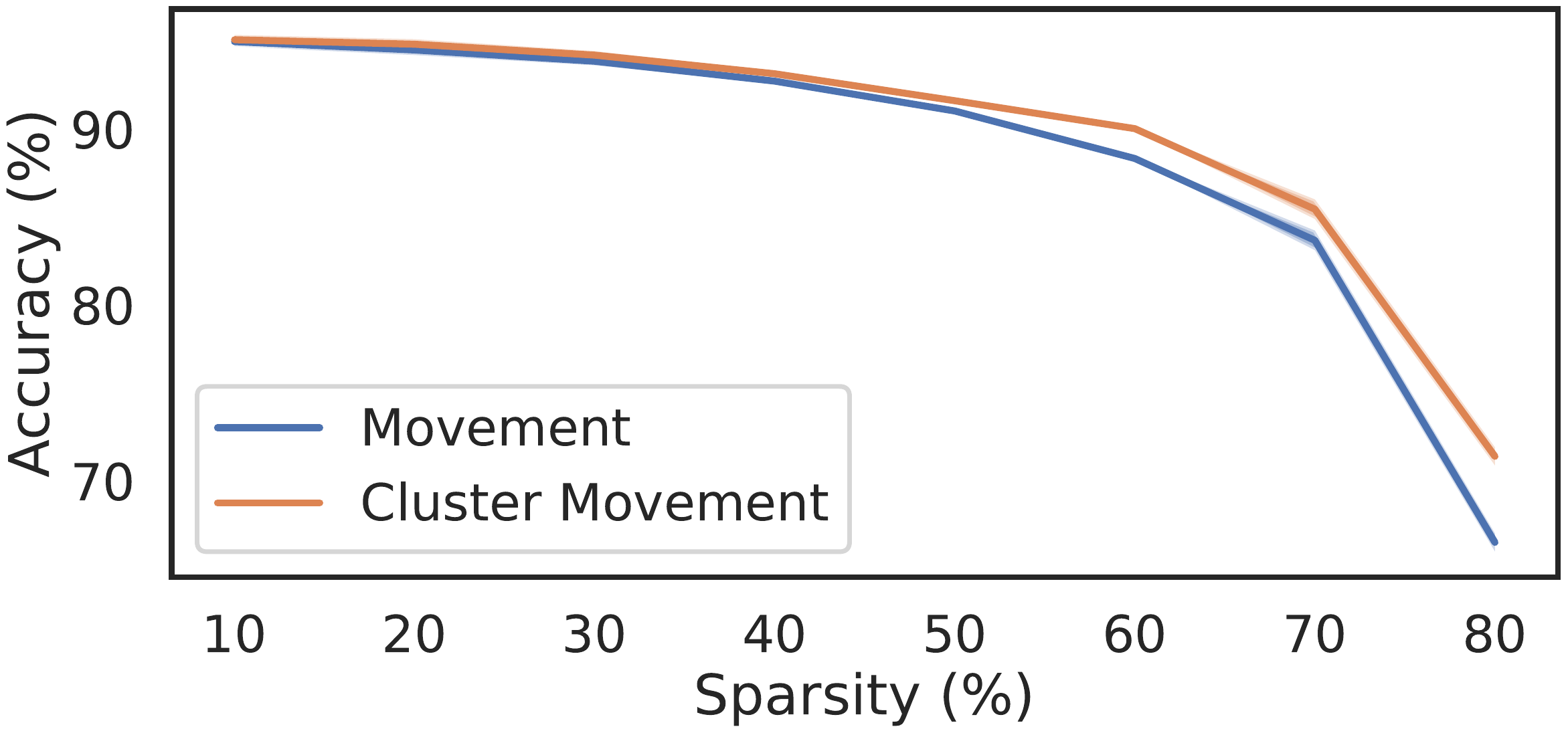}}
   \end{minipage}
   \caption{Results of the Lottery Ticket Hypothesis with Rewind test for different sparsities, performed with ResNet-18 on CALTECH-101.}

\label{LTH}
\end{figure}

\textbf{Comparison of Tickets. } We compare the performance of discovered tickets using the same pruning criteria, datasets and training procedure as described in Section \ref{Experiments}, and for the ResNet-18 architecture. As our experiment concerns large datasets and complex architecture, we propose to study the effect of our pruning technique on the Lottery Ticket Hypothesis with Rewinding. To uncover the tickets, we adopt the same methodology as presented in Section \ref{Methodology}, but reinitializing the weights after each pruning step to the value they had after the initial training step, \textit{i.e.} to their value after Step (1) in Figure \ref{process}. The operation is performed for each criteria evaluated in the Section \ref{Experiments} and for sparsity levels ranging from $10\%$ to $80\%$. After extracting the subnetwork, we train it for $15$ epochs and then compare the versions obtained with and without the addition of our variety enforcing clustering method. \\

\textbf{Results. } From this experiment, whose results are reported in Figures \ref{LTH},  we can observe that in most cases, the addition of the clustering method prior to the criteria selection helps to find a better performing ticket, thus validating the quality of the pruned network. While the addition of a clustering technique before applying the pruning criteria seems profitable in most cases, it benefits movement pruning the most. Indeed, increases up to $2\%$ in accuracy may be observed in the case of $l_1$ pruning, and up to $5\%$ in the case of movement pruning.

\section{Conclusion}

In this work, we propose a novel pruning method, introducing a clustering process before applying the pruning criteria. This clustering process, based on an interpretation technique called Activation Maximization, groups filters sensitive to similar features in the input image. By then applying the pruning criteria to each feature group, we ensure that pruning is applied on redundant filters, and that rare filters, which may be alone in their group, are retained. Experiments have shown that our method leads to better results than classical methods on both VGG-16 and ResNet-18 architectures and for CIFAR-10, CIFAR-100 and CALTECH-101 datasets. Those results demonstrate that one should avoid pruning rare or unique filters and that keeping a wide filter variability is crucial to achieving both a higher pruning rate and a lower accuracy loss. Moreover, by performing Lottery Ticket Hypothesis with Rewinding tests, we have demonstrated that the subnetworks discovered after pruning were of better quality, as they were able to reach higher performance in the same training time. 

%
%
%
%


\end{document}